\documentclass[preprint,12pt]{elsarticle}

\biboptions{sort&compress}
\usepackage[section]{placeins}
\usepackage{booktabs}
\usepackage{amssymb}

\journal{Additive Manufacturing}

\begin{document}

\begin{frontmatter}



\title{Deep-Learned Generators of Porosity Distributions Produced During Metal Additive Manufacturing}

\affiliation[inst1]{organization={Mechanical Engineering, Carnegie Mellon University},
            addressline={5000 Forbes Avenue}, 
            city={Pittsburgh},
            postcode={15213}, 
            state={PA},
            country={USA}}
\affiliation[inst2]{organization={Sandia National Laboratories},
            addressline={1515 Eubank Avenue}, 
            city={Albuquerque},
            postcode={87185}, 
            state={NM},
            country={USA}}
\affiliation[inst3]{organization={Chemical Engineering, Carnegie Mellon University},
            addressline={5000 Forbes Avenue}, 
            city={Pittsburgh},
            postcode={15213}, 
            state={PA},
            country={USA}}
\affiliation[inst4]{organization={Machine Learning, Carnegie Mellon University},
            addressline={5000 Forbes Avenue}, 
            city={Pittsburgh},
            postcode={15213}, 
            state={PA},
            country={USA}}

\author[inst1]{Odinakachukwu Francis Ogoke}

\author[inst2]{Kyle Johnson}
\author[inst2]{Michael Glinsky}
\author[inst2]{Chris Laursen}

\author[inst2]{Sharlotte Kramer}
\author[inst1,inst3, inst4]{Amir Barati Farimani}

\begin{abstract}
Laser Powder Bed Fusion has become a widely adopted method for metal Additive Manufacturing (AM) due to its ability to mass produce complex parts with increased local control. However, AM produced parts can be subject to undesirable porosity, negatively influencing the properties of printed components. Thus, controlling porosity is integral for creating effective parts. A precise understanding of the porosity distribution is crucial for accurately simulating potential fatigue and failure zones. Previous research on generating synthetic porous microstructures have succeeded in generating parts with high density, isotropic porosity distributions but are often inapplicable to cases with sparser, boundary-dependent pore distributions. Our work bridges this gap by providing a method that considers these constraints by deconstructing the generation problem into its constitutive parts. A framework is introduced that combines Generative Adversarial Networks with Mallat Scattering Transform-based autocorrelation methods to construct novel realizations of the individual pore geometries and surface roughness, then stochastically reconstruct them to form realizations of a porous printed part. The generated parts are compared to the existing experimental porosity distributions based on statistical and dimensional metrics, such as nearest neighbor distances, pore volumes, pore anisotropies and scattering transform based auto-correlations.

\end{abstract}




\end{frontmatter}


\section{Introduction}
\label{sec:intro}



    Additive Manufacturing (AM) is an iterative part construction process, which increases the feasibility of constructing complex three-dimensional parts. AM methods allow for the production of these parts while decreasing material waste, and enable both local control of material properties and the mass production of custom products \cite{li2020review, reeves2011additive}. Laser Powder Bed Fusion (L-PBF) is a subcategory of the broader class of AM methods, wherein a product is created by using a laser heat source to  produce cross- sections of a part by selectively melting the surface of a bed of metal powder particles. After the laser has completely melted a given layer, a thin layer of powder is redeposited on the surface of the powder bed. This melting and deposition process is repeated until the part fabrication process is completed. Powder Bed Fusion (PBF) methods have been used to construct products from a wide range of metallic alloys, and have seen heavy usage in the biomedical and aerospace industries \cite{cunningham2017analyzing,mower2016mechanical,spierings2013fatigue,lewandowski2016metal}. 
    The widespread adoption of these methods for precision use cases can often be challenged by the susceptibility of L-PBF parts to defects and inferior physical properties due to the nature of the build process \cite{kok2018anisotropy}. For instance, residual thermal stresses are often present in the material, in addition to surface roughness, porosity, delamination and cracking \cite{mani2015measurement,murr2012metal,al2010origin, markl2016multiscale}.

%

Detailed analyses of the mechanics leading to part failure have been carried out experimentally, using impact, fatigue, and stress tests to examine the effect of the microstructure on the structural integrity of the produced parts \cite{Vilardell2019, Andreau2019, Gong2015}. From these experiments, it has been shown that pore coalescence and void formation is a main driver of crack propagation, where crack propagation lowers the stress required for failure to occur \cite{Vilardell2019}. Additionally, Andreau et al. (2019) demonstrate that the distribution of pores in relation to each other, the pore proximity to the surface, and the surface roughness characteristics, are key determinants of the fatigue properties of manufactured parts \cite{Andreau2019}.  Therefore, microstructure characterization is necessary to analyze the potential fatigue behavior of a product fabricated using L-PBF.


Due to the influence of these microstructural features on the structural integrity of printed parts, methods have previously been developed for reconstruction and generation. These methods have been applied in related fields where physical behavior is dependent on microstructure properties, such as geosciences, electrochemistry, and materials science \cite{Siddique2012, Quiblier1984, Cnudde2013, Hasanabadi2016, Gerke2015}. For instance, Hasanabadi et al. (2016) introduced a method for reconstructing 3D realizations of isotropic and anisotropic microstructures for multiphase heterogeneous materials, using two-point correlation functions to ensure the statistical similarity of the 3D reconstruction to the original 2D cross-section \cite{Hasanabadi2016}. In a similar study, Gerke et al. (2015) propose a stochastic reconstruction technique to merge multi-scale images of shale rock, by combining a correlation function-based approach with simulated annealing in a compound model first proposed by Yeong-Torquato et al. (1998) \cite{Gerke2015, yeong1998reconstructing}. Using this technique, the authors demonstrate the ability to merge macroscale, microscale, and nanoscale features into a single image. 


%

While these techniques have successfully generated stochastic microstructures, the two-point probabilistic models they are based on can become computationally complex, and scale unfavorably with the size of the domain \cite{Moussaoui2018, Wang2018, Hasanabadi2016, Yao2013}. Therefore, deep learning methods have been introduced for manipulating data governed by high-dimensional probability correlations \cite{goodfellow2014generative, mirza2014conditional, farimani2017deep}. Specifically, deep generative models, such as the Generative Adversarial Network (GAN) architecture paradigm, have been used to rapidly construct stochastic examples of microstructure sections \cite{Mosser2017, Gayon-Lombardo, Janssens2020, Shams2020, Li2018}.

For instance, Janssens et al. (2020) make use of super-resolution GANs to upscale the resolution of existing microstructure samples for porous fluid flow, to alleviate the sample size versus resolution tradeoff. Using their GAN architecture, Janssens et al. are able to better resolve the pore network properties that determine the fluid flow behavior observed \cite{Janssens2020}. In related work, Gayon-Lombardo et al. (2019) produce super-resolved images of pore distributions by conditioning samples on low-resolution images.


This method is able to accurately generate new microstructure samples, however, the method relies on the use of periodic boundary conditions to limit the computational size of the model \cite{Gayon-Lombardo}. However, it is difficult to define periodic boundary in cases where the position of the 3D boundary directly influences the porosity distribution. This is also the case for where the size and porosity of the part are not well-suited for selecting meaningful three-dimensional samples by segmenting the image into equally sized sub-volumes. Therefore, this motivates the use of a model that separates the large scale generation process from the small scale generation process for computational feasibility.  In this work, we separate the generation process methodology based on scale, with probabilistic sampling used at large scales to generate porosity distributions, GANs used at the intermediate scale to generate pores, and scattering transformation based tools used at the smallest scale to generate surface roughness.

The Mallat Scattering Transform (MST) is a physics-based analogue to convolutional neural networks that stores statistical information and two-point correlations in a reduced order format. These coefficients have been used previously for machine learning based classification, due to their ability to extract the multi-scale structure of the data \cite{saydjari2021classification, cheng2020new, glinsky2020quantification}. Due to these properties, the MST can be used as a tool for generating small-scale features with limited data. For instance,  Saydjari et al. (2021) leveraged the ability of the MST to extract information from large-scale structure in magnetohydrodynamic simulations \cite{saydjari2021classification}. In their work, a classification task is performed on the coefficients produced by the transform, yielding the key hydrodynamic parameters associated with the simulation from the behavior of dust density fields. In another application, Glinsky et al. (2020) made use of the MST to develop a classification system for the morphology of plasma signatures during magnetized nuclear fusion \cite{glinsky2020quantification}.

This work presents a method for stochastic reconstruction of novel porosity distributions in 3D AM-produced parts, by separating the process into the pore generation, surface roughness generation, and pore placement sub-problems. By doing so, we are able to address the generation process at multiple scales, leveraging more computationally expensive correlation-based methods for small scale feature generation, and using coarser probability sampling to create the global porosity distribution.  For the pore generation problem, we use 3D Generative Adversarial Networks to reconstruct novel examples of individual pores that globally approximate the distribution observed in the original dataset. To create novel examples of the surface roughness, we implement a microcanonical model based on the Mallat Scattering Transform (MST) that learns the features of the random process associated with the profilometry extracted \cite{zhang2021maximum} via a set of scattering coefficients.

Finally, we reconstruct these components into an overall part by combining the surface roughness and pores by sampling from the established spatial distributions of pore properties. This method preserves the multi-scale statistical relationships present in the original part that determine its mechanical properties, and allows for explainable generation of an arbitrary number of samples from a relatively small amount of tomography data. In Section \ref{sec:methods}, we present the methodology and considerations for this technique. In Section \ref{sec:results}, we discuss the results obtained by applying this method to L-PBF-produced Al-10Si-Mg tensile coupon samples. Additionally, the successful implementation of this method allows for the creation of unique, accurate porosity initializations for finite element analysis, enhancing the understanding of the process-structure-property relationships, while bypassing the cost of printing samples and performing destructive testing.


\section{Methodology}
\label{sec:methods}

\begin{figure*}[!htbp]
\begin{center}
   \includegraphics[width=1\linewidth]{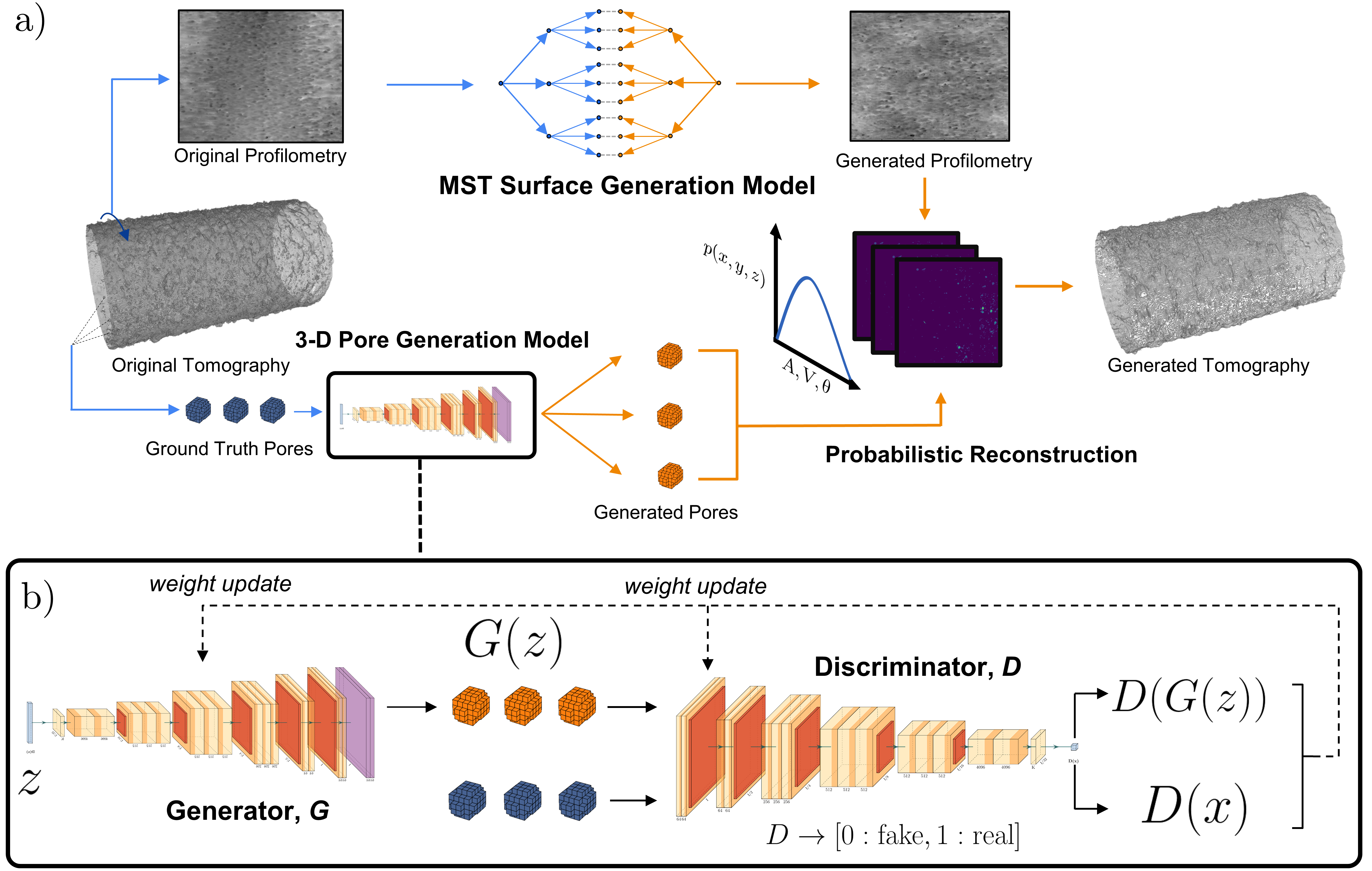}
\end{center}
   \caption{a) The overall pipeline of the process, from the ground truth set of tomographies, to the generated part reconstruction. b) The 3D DCGAN used to generate novel examples of individual pore samples. A generator network $G$, and a discriminator network $D$, are jointly trained adversarially against each other. The discriminator network aims to determine if a sample is real, i.e., if it is a member of the ground truth dataset. The generator aims to produce novel synthetic samples that are similar to the ground truth, such that the discriminator predicts that the synthetic samples are real samples. }

\label{schematic}
\end{figure*}

\subsection{Deconstruction}

In the overall pipeline for this work (Figure \ref{schematic}), the 3D CT scan data is first analyzed to extract the pore distribution and surface contour. A dataset of individual pores is formed from the pore distribution, saving each individual pore’s three-dimensional shape as a group of connected voxels. A dataset of surface contours is also formed, by first transforming the approximately cylindrical boundary into polar coordinates, and then calculating the deviation from a perfect cylinder as a function of the axial distance, $z$, and angle, $\theta$. Following this, a new realization of surface roughness is generated. Using this generated surface, the probability distribution of pore metrics with respect to the $(x, y, z)$ location inside the part is used to sample a combination of pore metrics that specifies a pore morphology. Given this pore specification, which contains information about the desired volume, orientation and shape of the pore at hand, a new pore is sampled and placed within the domain at the location indicated. Finally, post-processing is carried out to account for the interaction of the surface boundary with the placed pores. 

\subsection{Pore metrics}
The properties of each pore in the sample are extracted and analyzed to ensure the porosity distribution in the generated part is consistent with those found in the ground truth samples.  The volume of the pore is computed based on the binarized 3D pore representation. For additional spatial metrics, the moment of inertia is used to extract information on the shape and positioning of the pore. The moment of inertia tensor is first calculated from the 3D pore representation using  the computed three-dimensional image moments. This tensor is then decomposed into the corresponding eigenvalues and eigenvectors for further analysis. The eigenvalues of the matrix are then used to calculate the anisotropy, $A$, of the pore according to Equation \ref{equation:anisotropy}, where $\lambda$ is the set of eigenvalues of the moment of inertia tensor.

\begin{equation}
\label{equation:anisotropy}
    A =  1 - \frac{\lambda_{min}}{\lambda_{max}}
\end{equation}

The orientation of the pore is extracted by calculating the angle between the principal eigenvector of the pore, and the unit vector corresponding to the tensile axis, $z$. The angle $\phi$, from this principal component to the x-axis is also found to measure the angle of the pore’s orientation in the x-y plane. In addition to these metrics, the nearest-neighbor distance distribution is also computed, as this correlates with the tendency of multiple closely located pores to coalesce and nucleate cracks in the material. The nearest neighbor distance is found as the minimum over the  set of distances from a pore's centroid to all other pore centroids in the segment.

\subsection{MST Microcanonical model for Boundary Generation}

The Mallat Scattering Transform (MST) is a cascade of wavelet transformations with non-linear smoothing operators that act as an explainable analogue to deep convolutional neural networks \cite{mallat2012group, mallat2018phase}. The MST relies on fixed convolutional filters generated from rotating and scaling a mother wavelet, $\psi$.  In the MST process, an input signal is sequentially convolved with a set of these mother wavelets at different scales and rotational phases, and the output of the convolution cascade is then convolved with a father wavelet, $\phi$ to produce a set of coefficients for each convolution layer \cite{mallat2012group}.

The mother wavelet family of convolution filters is defined in terms of scale and phase as

\begin{equation}\label{equation:motherwavelet}
\psi_{2^j r} (u) = 2^{2j}\psi(2^jr^{-1}u)
\end{equation}

where $j$ indicates the scale of the wavelet filter, $u$ is the space on which the wavelet is defined, and $r$ is the rotation of the wavelet. The output of this convolution is then convolved with the spatial averaging father wavelet, which is analogous to a pooling operation. The father wavelet, $\phi_J$, is specified as
\begin{equation}
  \phi_J = \phi_{2^J} (u) = 2^{-2J}\phi(2^{-J}u)  
\end{equation}

where $J$ parameterizes the maximum scale of the transformation.

The first and second order transform coefficients, $S_1[p]x$ and $S_2[p,p']x$, are  given by the following respectively:
\begin{equation}
    S_1[p]x =  \left | x * \psi_{\lambda_1}* \phi_J\right |
\end{equation}

\begin{equation}
    S_2[p,p']x =  \left | | x * \psi_{\lambda_1}* \phi_J\right | * \psi_{\lambda_2}|*\psi_J
\end{equation}

where $p$ and $p'$ refer to the scale-dependent path of a scattering coefficient at each order. Here, $\lambda_1$ and $\lambda_2$ refer to the scale and rotation of the mother wavelet used at each order based on the expression $\lambda = 2^j r$, and $S[p]x$ are the set of scattering coefficients. $J$ first order coefficients  and $\frac{J(J-1)}{2}$ second order coefficients are produced, where $J$ is a hyperparameter that controls the size of the father wavelet and the scales of the generated mother wavelets. The combined coefficients of the first and second-order MST transform have been shown to be a  statistically complete metric, preserving 98 - 99 \% of the energy from the original signal based on the value of $J$ \cite{mallat2012group}. This renders higher-order transforms unnecessary.

During the pre-processing stage, the cross-section of the part boundary is extracted at each $\Delta z$ interval along the tensile axis, and converted to polar coordinates. This conversion yields a line plot of the radius, $r$ of the part boundary, as a function of $\theta$. These profiles are then extracted and stacked for multiple cross-sections along the axial direction to form an image, as seen in Figure \ref{deconstruction}. Using this image, a new realization of the surface roughness can be generated using models that identify the underlying random process.


To make new examples of these contours, we use a gradient-based microcanonical model, described in more detail in \cite{zhang2021maximum}. This model operates by creating an image with the same statistical properties as the ground truth image, with the statistical metric defined by the covariance of the MST coefficients. As the MST operates by performing iterative wavelet transforms at multiple scales, it provides a set of coefficients that encodes the information in the uncompressed image, as described by \cite{mallat2012group}. Here, an image  initialized to white noise undergoes gradient descent to approach an image with the same properties as the ground truth, as defined by the covariance of their respective MST coefficients. The loss function is given in Equation \ref{equation:mstloss}:

\begin{equation}\label{equation:mstloss}
    f(x) = \mathbb{E}_G || \tilde{K}_{Rx} - \tilde{K}_{R\bar{x}} ||^2
\end{equation}
where $R$ is the MST operator, $\tilde{K}_{Rx}$ is the covariance of the MST coefficients from the optimized image, and $\tilde{K}_{R\bar{x}}$ is the covariance of the MST coefficients of the ground truth. An ensemble of $G$ different MST transformations are created to introduce stochasticity and construct multiple distinct realizations from a single image. This ensemble of transformations is generally constructed by applying various operations to the target image, such as translations and rotations, effectively performing data augmentation. In this work, we construct the ensemble of $G$ samples by performing periodic translations on the original image.


During the training process, the initial image is first resized to a smaller 256$\times$256 image to limit memory usage while maintaining image quality. Following this, the intensity means along both axes of the image are removed. From these pre-processed images, the microcanonical model is trained to generate new realizations. Once generated, a Savitsky-Golay filter \cite{savitzky1964smoothing}, with a window size of 100 $\mu m$ and a polynomial order of 4, is applied to these realizations to eliminate sharp edges and discontinuities. Finally, these examples can be transformed back into three dimensions, by converting each line segment of the image back to polar coordinates to form a 2D cross-section, and stacking these 2D cross-sections in the direction of the tensile axis.



\subsection{Sampling method}
To create the large scale porosity distribution, a stochastic model is developed by constructing a conditional probability distribution for the pore properties, based on their spatial location. Generally, a distribution $P(x,y, k)$ is developed for each pore property $k$, to sample pores that match the original spatial distribution. These probability distributions are constructed with considerations for computational expense by binning pores into larger windows of size $\frac{M}{N_{b}}$, where $M$ is the diameter of the part sample and $N_{b}$ is the number of equally spaced bins used to construct the probability estimates. 

The properties chosen are orientation, anisotropy, volume, and the number of pores found in a certain  bin. These specific properties are chosen to characterize each of the pores, as they are shown to influence the final properties of the part following processing \cite{salarian2020pore}. The probability distributions for these variables are selected by binning the pores according to their position in the $(x,y)$ plane, and developing a custom distribution for these properties accordingly. The resulting $N_{b}\times N_{b} \times N_{b}$ matrix is used for reconstruction, with $N_{b}$ becoming a tunable hyperparameter. As $N_{b}$ increases, the stochasticity is reduced, as the probabilities enforced by each bin become increasingly strict. However, as $N_{b}$ decreases, there is an increasing opportunity for random distributions to occur, at the cost of diverging from the original property distribution. The qualitative influence of this parameter is discussed in more detail in \ref{sec:sample:appendix}.

During traversal, the probability distribution of each property is used to sample a specific number of pores. The number of pores, $N_{p}$, is computed based on the density distribution of pores with respect to their $(x,y)$ location, extracted earlier from the ground truth experimental parts. Following this process, the volume, anisotropy, and orientation values are then sampled $N_{p}$ times from the interpolated distributions that are defined by their occurrence in the ground truth dataset. Next, the locations of the pore centroids are calculated by sampling a uniform distribution within in the window boundaries, assuming the properties are locally statistically stationary.
The advantages of this model approach, when compared to a purely GAN based microstructure reconstruction, lies in the fact that it enables customization of the degree of stochasticity of the reconstruction though the $N_{b}$ parameter. It  also provides a more interpretable model by allowing the user to directly view and modify the spatial probability distributions used for sampling.

Finally, the separation of the sub-problems encodes an explicit check of the ML produced pores to ensure they are physically accurate. For instance, a GAN may undergo cases of mode collapse during the prediction task due to the complexity of the parameters involved and the lack of interpretable training metrics, as demonstrated in \cite{goodfellow2014generative}. In the production scenario, where many parts are created in sequence, a silent failure in a purely GAN-based pipeline would not be caught in an automated process unless each instance is independently analyzed. Here, we analyze each pore during the embedding process via the probability based sampling process, which selects the pore among the general bank of pores that is closest to the prescribed parameters. This prevents generated pores that are unrealistic from being embedded in the produced synthetic samples.

\subsection{GAN models}

In the Generative Adversarial Network (GAN) architecture, first introduced in Goodfellow et al. (2014) \cite{goodfellow2014generative}, two competing networks, a discriminator $D$ and generator $G$, produce novel examples of an input given a latent noise vector, $z$. The discriminator network aims to be able to distinguish real samples from fake samples, while the generator aims to produce fake samples that are realistic enough to cause the discriminator to believe that they are real samples. The overall architecture of this process is presented in Figure \ref{schematic}b). The two networks are then optimized with a game-theoretic loss function, $L$, that aims to reach a Nash equilibrium between the two objectives.

\begin{equation}\label{equation:ganloss}
    L = E_x [log D(x)] + E_z[log (1 - D(G(z))]
\end{equation}
The discriminator network, $D$, outputs the confidence that a given sample is real  as a value between $0$, where it believes the sample is synthetic, and $1$, where it believes the sample is a real example from the ground truth training data. The generator network acts to minimize the loss function above by producing realistic samples, i.e., where $D(G(z))=1$. Conversely, the discriminator acts to maximize the function by correctly predicting $D(x)=1$, and $D(G(z))=0$, where $x$ is the ground truth data, and $G(z)$ is a generated sample. 

\subsection{3D GAN for Pore Generation}

  Given the generated pore metrics sampled from the probability distributions, a three-dimensional GAN is used to create new pore objects. The architecture used here is based on the three-dimensional Deep Convolutional GAN architecture developed by Wu et al. (2016) \cite{wu2016learning}. During the generation process, a randomly sampled 100-dimensional latent vector $z$ is passed to a generator network, which will create a new $2\times64\times64\times64$ tensor, where each of the two channels represents the probability of a given voxel either containing pore material or solid phase. This tensor is provided to the discriminator network, $D$, which outputs the confidence that the sample corresponds to the ground truth distribution.
  
  A single pore can be extracted from the output of the generator network, by creating a binary image based on the probability of a porous phase occurring in a voxel. Therefore, this 3D GAN model is then used to generate a bank of 50 000 pores, which are then analyzed to ensure the bulk properties of the produced pores aligns with the bulk properties of the ground truth data distribution, and sampled during the overall component reconstruction process. The statistical properties for a subset of 1200 pores are shown in Figure \ref{ganresults}. 
  


\subsection{Reconstruction}

To avoid placing pores in conflicting locations where they may overlap, each pore is checked before insertion into the overall part. This check determines whether the addition of the pore to the proposed location  will result in exactly one more pore being added to the local surrounding solid phase. If this pore were to overlap with already present pores, the total number of pores would remain constant, or decrease. This prevents accidental pore recombination.


Due to the memory usage scaling with the size of the component, a stitching method is also implemented to allow for piecewise analysis. First, a moving window of size $\Delta z$ is defined for traversal of the part through the tensile $z$-axis. During this process, an initial boundary frame of the window $z_0$ is selected as the left-hand side of the moving window, and an ending frame, $z_f$ is chosen as the final boundary for this moving window. The pores in contact with this final ending frame are removed, as they are bisected by this frame and would otherwise cause partial pores to be included in the analysis, distorting the anisotropy and orientation measurements. The window is then advanced by $\frac{\Delta z}{2}$, half of the window size, to cause the area currently being examined to overlap with the previous area by this amount. During this process, pores that were previously bisected by the end boundary in the previous moving window are now considered in the calculations, while pores that were entirely encapsulated by the previous boundary are removed from the analysis. In this manner, a continuously moving window can store and process the thousands of pores observed in arbitrarily large part samples, avoiding escalating memory requirements. 

\section{Results}

The initial data consists of twelve 3D computed tomograph samples of Al10SiMg, at an average resolution of 4 $\mu m$ per voxel length. Each tomograph consists of 3000 segmented cross sections of the part, with an average resolution of 600 $\times$ 600 pixels. The average diameter of each sample is 2.4 mm, and the average length along the tensile axis is 14.6 mm.

\subsection{Boundary Generation}

\begin{figure*}[!htbp]
\begin{center}
   \includegraphics[width=1\linewidth]{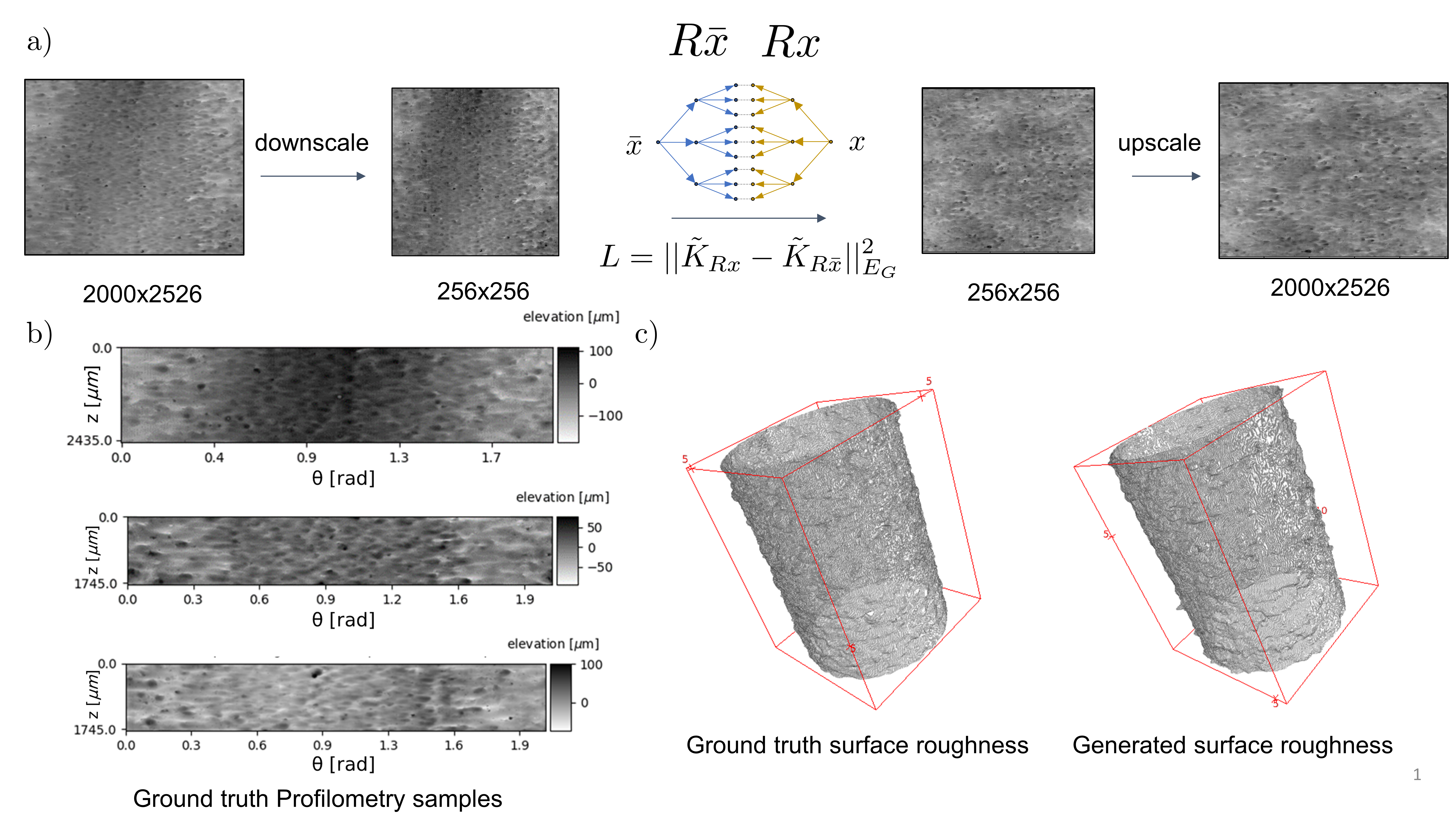}
\end{center}
   \caption{A schematic of the microcanonical model used to generate new part boundary examples. The covariance of the MST coefficients from a ground truth surface roughness profile is compared to the covariance of the MST coefficients of an optimized trial image. a) In order to create new realizations of the surface roughness profilometry, the extracted surface is first downsampled to a coarse resolution of $256 px \times 256 px$.  An ensemble of Mallat Scattering Transforms (MSTs) are applied to this downsampled image, and the covariance of the produced coefficients are used to generate a new realization, following the method first proposed in \cite{zhang2021maximum}.  Specifically, an initially white noise image undergoes gradient descent to produce an image with the same MST covariance as the ground truth image, minimizing the loss function  $f(x) = || \tilde{K}_{Rx} - \tilde{K}_{R\bar{x}} ||^2_{E_G}  $. Here, $R$ is the MST operator, $\tilde{K}_{R\bar{x}}$ represents the ground truth coefficient covariance, and $\tilde{K}_{Rx}$ is the trial image. b) Example profiles produced from the experimental dataset, showing the surface roughness projections for three different parts. c) A comparison of the three dimensional ground truth surface roughness to the generated surface roughness } 

\label{deconstruction}
\end{figure*}

\begin{figure*}[!htbp]
\begin{center}
   \includegraphics[width=1\linewidth]{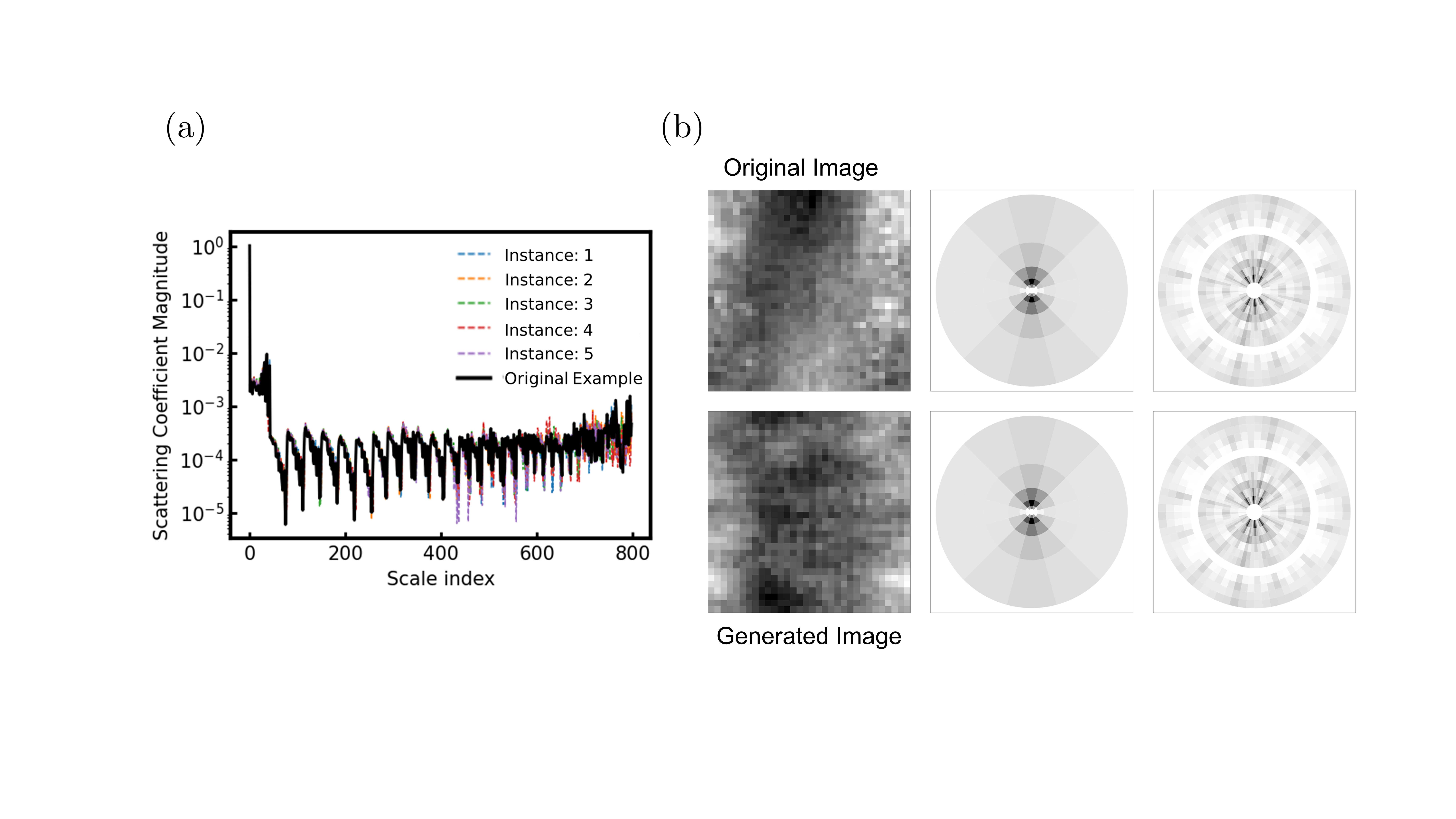}
\end{center}
   \caption{The generated images can be compared to the ground truth image by examining the produced Mallat Scattering Transform (MST) coefficients. These coefficients vary in phase and scale, and thus can be represented on a rose plot with scale changing along the radial axis. (a) The produced MST coefficients arranged in terms of scale for five different instances. The large scale structure at high scale values are preserved exactly, while the small structures vary across instances, introducing stochasticity to the realizations. b) Rose plots of the produced MST coefficients for both the original surface roughness profilometry sample (top) to the generated case (below) reveal very similar distributions of the coefficients with respect to angle and scale. $1^{\circ}$ refers to the first-order MST, while $2^{\circ}$ refers to the second-order MST.
}

\label{mst}
\end{figure*}

The first step in the reconstruction process is to construct the boundary enclosing the finished part. As the roughness of the surface will play a critical role in determining the fatigue strength of the material, we attempt to replicate the statistics of the original surface roughness boundary closely. 
The microcanonical model is trained for 500 iterations, using an MST model with four scale-based subdivisions and four phase-based subdivisions. 
Following a study on the reconstruction similarity compared to the number of iterations, the training process is capped at 500 iterations to reduce computational expense, as any improvements after 500 iterations are relatively marginal. 
Following training, the extracted samples are first qualitatively compared from the ground truth to the reconstructed cases. From this analysis, it can be seen that the degree of fluctuation of the surface roughness is visually similar between the cases. This is presented in Figure \ref{deconstruction}, as a comparison of the two unrolled boundary images. A more detailed quantitative comparison is also carried out with respect to the MST coefficients produced from both images. The MST provides scale-dependent coefficients describing the structure of a signal, analogous to a convolutional neural network with fixed weights. Therefore, examining the produced coefficients can provide a measure of the similarity between the two images. In Figure \ref{mst}, the distribution across five sample reconstructions is shown, with ground truth surface roughness scattering coefficients shown in black, and the instances shown as line plots surrounding the initial image. The large scale coefficients, closer to index 0, exhibit a close agreement to the ground truth image, demonstrating that the large scale structures of the surface image are preserved from the ground truth image to the stochastic reconstructions of the surface roughness. However, there is a larger variance band for the small scale coefficients, indicating that the model is able to vary the small-scale, specific features of the surface roughness to create unique realizations as the large scale structure remains intact. 

\subsection{3D Pore Generation}

\begin{figure*}[!htbp]
\begin{center}
  \includegraphics[width=1\linewidth]{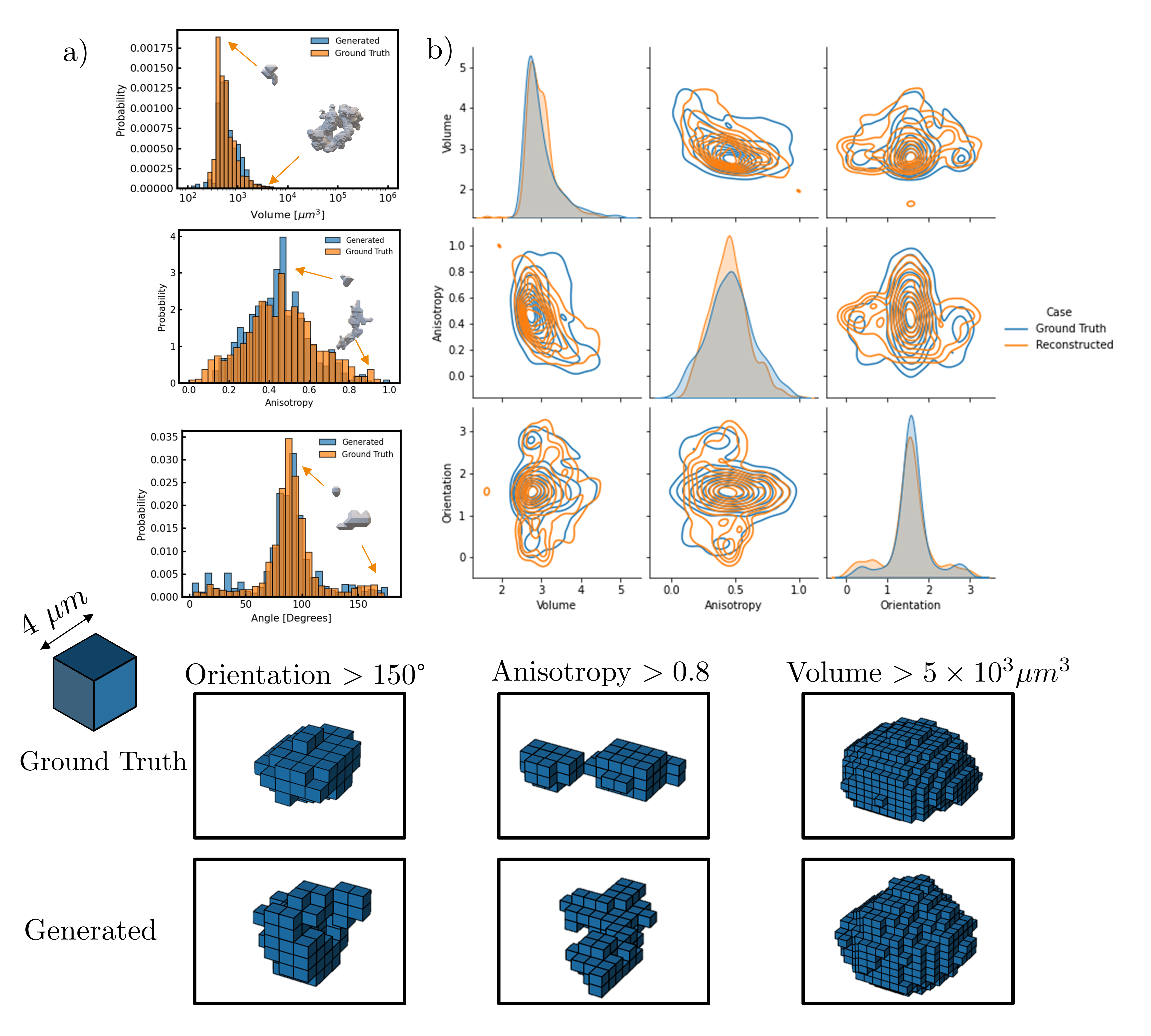}
\end{center}
  \caption{Statistics for 1200 generated parts using the 3D-DCGAN model. a) Probability Density Functions of the individual properties reveal agreement between the generated pores and the original, ground-truth pores.  Additionally, in a) sample pores from the ground truth experimental data, at various points inside the distribution, are displayed. b) Cross-correlation kernel density estimates are shown for the ground truth and generated pores, demonstrating agreement between the bivariate distributions generated by the model, and those found in the input dataset.}

\label{ganresults}

\end{figure*}

A 3D Deep Convolutional GAN is used to create novel realizations of the individual pores contributing to the porosity of the printed part. The GAN model with the architecture described in \ref{sec:hyperparams} is trained for 40 epochs on a dataset of 165,000 pore samples. These pore samples are extracted from all twelve computed tomography samples of the printed microstructure, isolated, and placed at the center of a 64$\times$64$\times$64 domain to create a standardized spatial positioning for training. Following training, the generated pores are analyzed to examine the proximity of their dimensional metrics to the distributions observed in the original dataset. Figure \ref{ganresults} demonstrates the agreement of the aggregate orientation, anisotropy, and volume distributions for the generated pores, relative to the ground truth pore samples extracted directly from the original part microstructures.

While we demonstrate that the aggregate statistics for the generated pores match the original distributions, the correlation between individual metrics of pore behavior must also be studied to ensure that the implicit relationships between these properties are preserved. For instance, while the independent population of large pores and the population of spherical pores may be the same across the generated and ground truth samples, the population of pores that are both large and spherical will also influence the properties of the regenerated microstructure. To examine these relationships in further detail, we produce kernel density estimation plots for each pair of metrics studied. In Figure \ref{ganresults}, the kernel density distributions between the ground truth and the reconstructed cases are well-aligned, indicating that stochastic realizations of new pores from the GAN model agree with the bivariate correlations seen in the data.


\begin{figure*}[!htbp]
\begin{center}
   \includegraphics[width=1\linewidth]{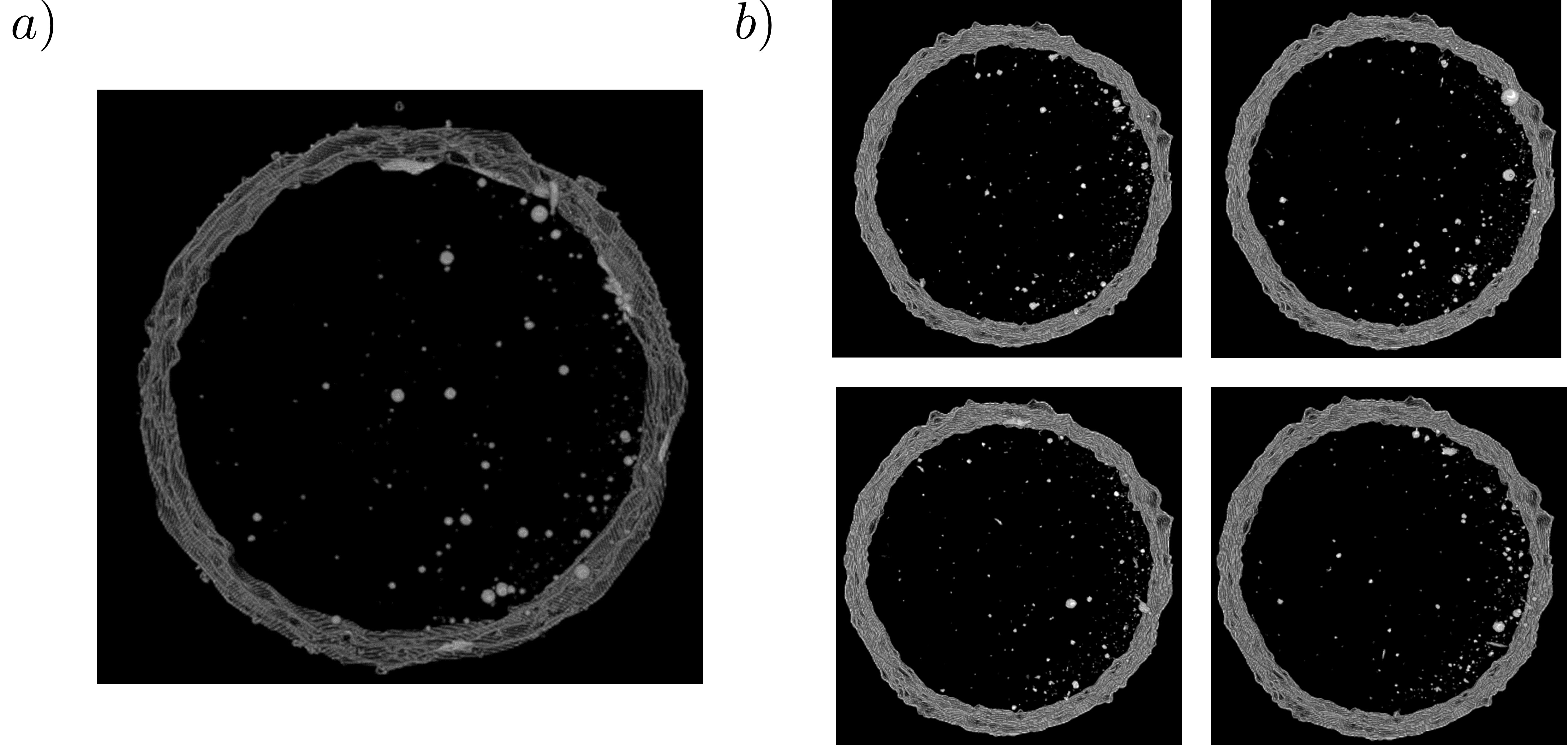}
\end{center}
   \caption{(a) A 35 $\mu m$ long segment of a part sample, oriented along the tensile axis. (b) Four realizations, also 35 $\mu m$ in length, using the generation pipeline to reconstruct new instances of the porosity distribution. The generated samples exhibit similar spatial and volumetric distributions when compared to the ground truth. Notably, the concentration of pores is consistently higher on the right side of the sample in both cases, and the size distribution of the produced pores appears visually similar.}

\label{generatedtomograph}
\end{figure*}

\begin{figure*}[!htbp]
\begin{center}
   \includegraphics[width=1\linewidth]{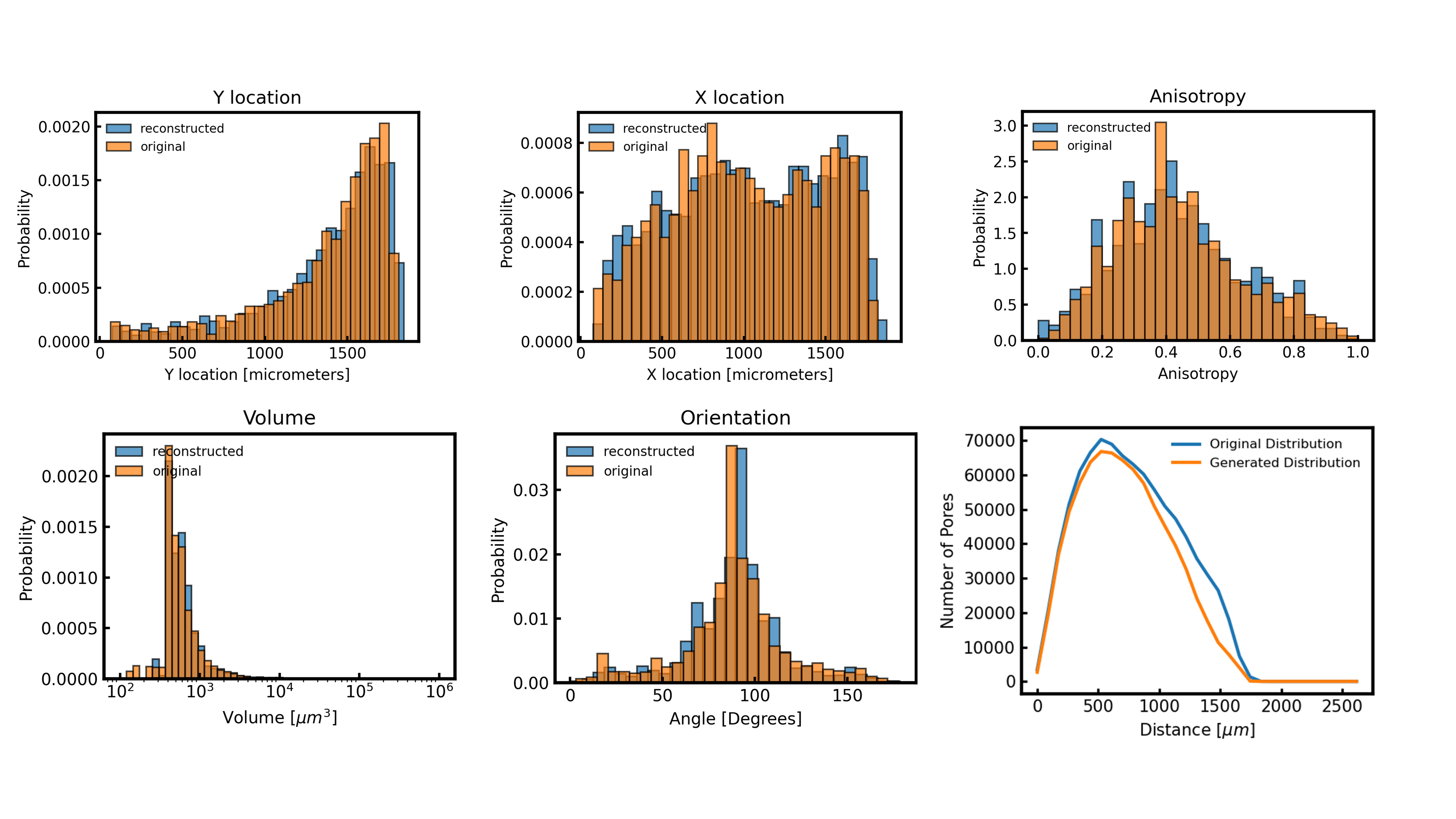}
\end{center}
   \caption{Statistics for a 7-mm segment of a generated part sample, oriented along the tensile axis. The orientation, locations, volumes and anisotropy are used to gauge the  agreement between the generated instance and ground truth examples. The nearest neighbor distance distribution of the pores in the reconstructed part is also used to gauge the performance of the model. Agreement is observed across all six examined metrics.}

\label{individualstatistics}
\end{figure*}

\subsection{Reconstructed Parts}
\begin{figure*}[!htbp]
\begin{center}
    \includegraphics[width=1\linewidth]{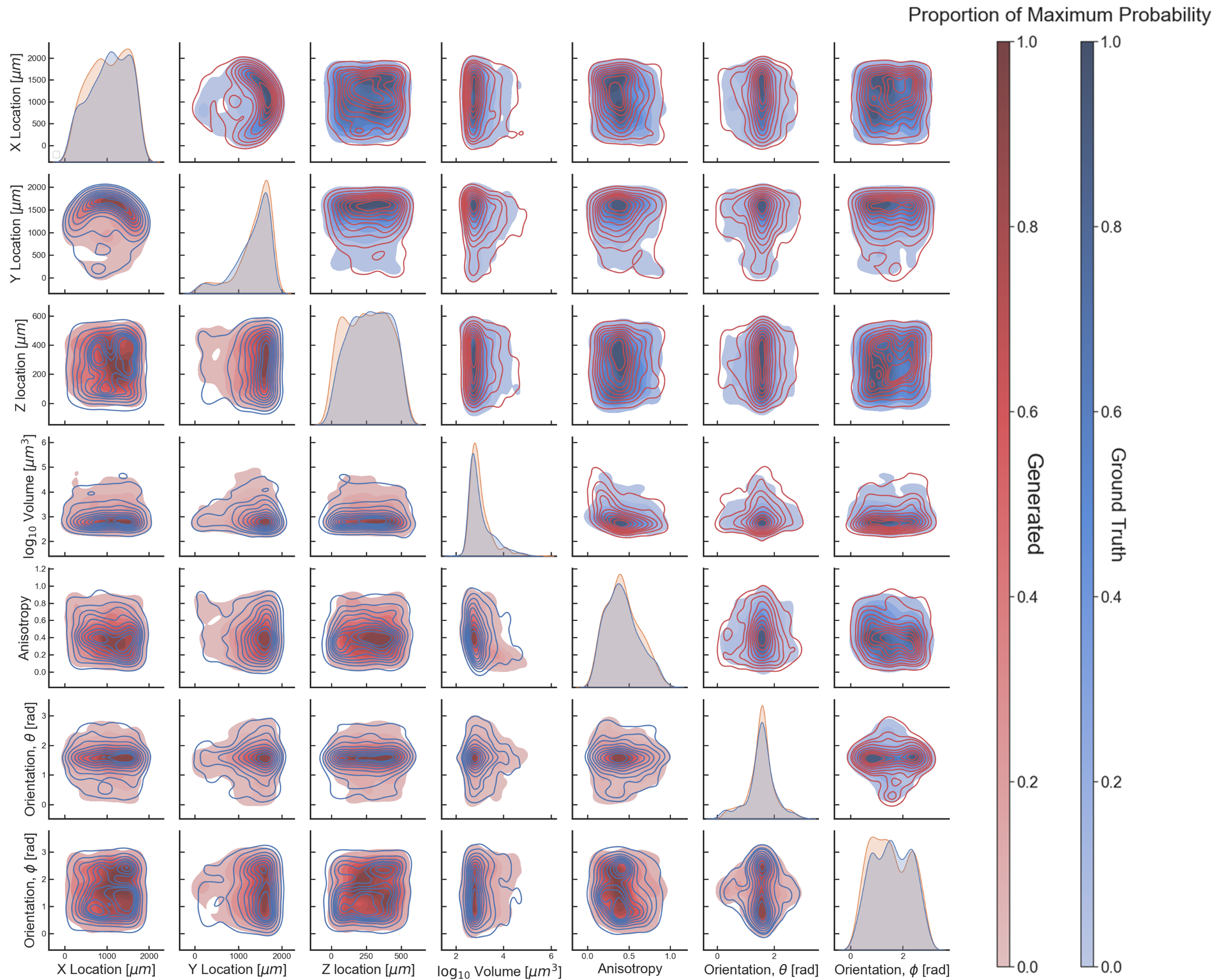}
\end{center}
   \caption{Cross correlation pair plots of the extracted pore properties reveal agreement between the trends observed in the ground truth data, and the trends observed in the reconstructed data. The bivariate correlations between extracted metrics are preserved from the original data, to the generated data. The individual levels on the contour line plots represent the probability distribution from 5\% to 95\% of the maximum probability, in intervals of 10\% between each level.}
\label{pairplot}
\end{figure*}

\label{sec:results}
The output of four instances of the generated sample, with a tensile length of 35 $\mu m$, is shown in Figure \ref{generatedtomograph}. When compared to the ground truth image, the four samples shown replicate the same distribution of porosity within the original sample, and exhibit variation in the exact location of pores from sample to sample. There is a large concentration of pores on the right side of the part, and relatively few pores on the left-hand side of the part sample. In both cases, one can also observe the presence of pores lying along the plane of the boundary, such as in the bottom left plot. Also, the volumetric distribution of the original pores is similar visually, where a relatively small number of large pores are observed, and occurrence increases as the pore volume decreases.

In order to verify that the pore distribution generated from the sample matches the ground truth, we extract the three-dimensional properties of the reconstruction, and compare them to the properties found in the original sample. To benchmark the quality of the reconstruction, the 2D Cartesian coordinates $(x,y)$, the anisotropy, and the orientation of the pore are calculated over a 7-mm segment of reconstructed part, and compared to the properties found in the original case. In Figure \ref{individualstatistics}, it can be seen that suitable agreement is observed for all of these quantities, as the two distributions are very similar for each of the properties observed.

In the anisotropy figure, the reconstructed version preserves the peak anisotropy occurrence at 0.5, indicating that the majority of the pores are non-spherical with a principal dimension twice the size of the smallest dimension. The orientation distribution is also preserved as well. The orientation distribution peaks at an angle of 90 degrees from the tensile axis, indicating that most pores have a principal axis that is oriented perpendicularly to the build direction. This indicates that pores are often formed from Lack of Fusion defects during the melting stage. The volume distribution also provides an insight into the distribution of pores. In both the reconstruction and the original part, there is generally an increasing amount of pores as size decreases, up to a threshold of 2700 $\mu m^3$, below which they cannot be reliably identified. This identification limit is due to the size threshold, which requires at least 8 contiguous voxels to be present for an object to be classified as a pore. This is imposed during pre-processing to avoid contaminating the statistics with incompletely resolved objects. 

\begin{figure*}[!htbp]
\begin{center}
   \includegraphics[width=0.9\linewidth]{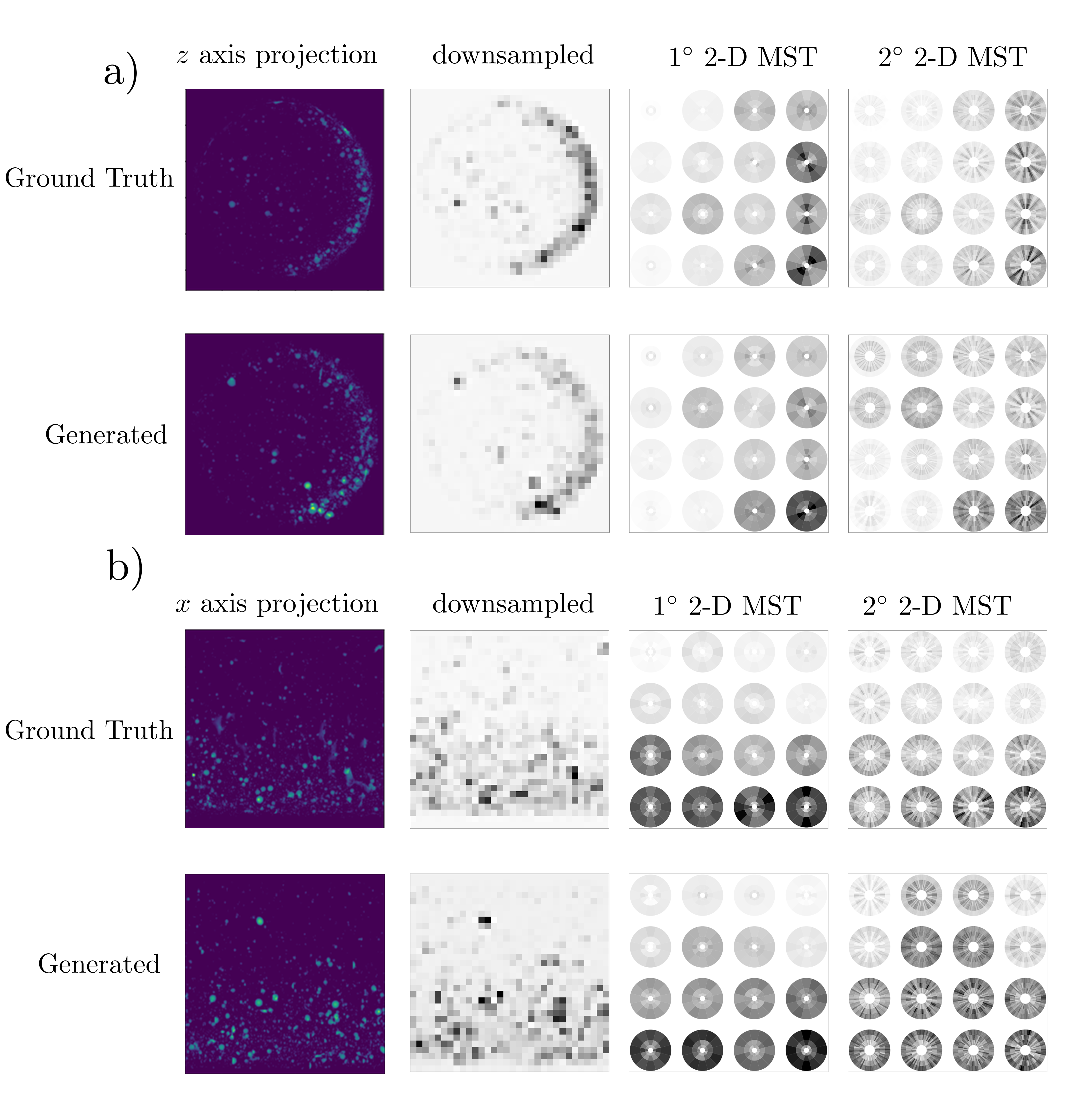}
\end{center}
   \caption{The Mallat Scattering Transform (MST) is taken in 2D on both the $z$-axis projection (a) and the $x$-axis projection of the 3D cylindrical part (b), for existing and new realizations. General agreement in terms of the overall structure is observed, while the fine scale structure varies. The 2D MST is used as opposed to the 3D MST, to reduce the memory required for computation. $1^{\circ}$ refers to the first-order MST, while $2^{\circ}$ refers to the second-order MST.}

\label{mst_projections}
\end{figure*}

To verify the statistical agreement between the generated and reconstructed samples, the 2D MST is performed on both the generated and reconstructed parts, following a projection onto a spatial axis. By construction, the MST is established as a near-unitary operator that conserves statistical information between the signal and the produced coefficients\cite{mallat2012group}. Mallat et al. demonstrate the completeness of the metric by describing the ability of the transform to encode information relating to 2D stochastic processes, which makes it suitable for quantifying the similarity between the generated and reconstructed samples.\cite{mallat2012group}.   Therefore, examining the MST can provide insight into the information preserved during the reconstruction process. The results of this process are shown in Figure \ref{mst_projections}, demonstrating a general agreement between the large-scale structures of the porosity distribution. The ability to exactly reconstruct the MST of the ground truth sample in the generated case is limited by the stochasticity of the reconstruction, as well as the computational expense of using the MST as a direct metric for the generation of new samples. For computational efficiency, the 2D MST on a projection of the component is used instead of the 3D MST on the entire component. 

The quality of the reconstruction precision can be defined by examining the proximity of the MST coefficients of the generated image to the original image. Specifically, we define a precision metrics on the ground truth and generated samples, $P$ and $\hat{P}$ respectively, which measures the variation of the MST coefficients between different samples. This effectively measures the radius of the distribution of the experimental samples as $P$, the radius of the distribution of the newly constructed realizations as $\hat{P}$.

We can define $P$ and $\hat{P}$ as  

\begin{equation}\label{equation:precision}
   P = \frac{\mathbb{E} \left ( \left \| SX - SX \right \|^2 \right )}{2 \mathbb{E} \left ( \left \| SX \right \|^2 \right )} = \frac{ \mathbb{E}\left [ \left\| SX \right\|^2 \right ] - \left\| \mathbb{E} \left [ SX \right ] \right\|^2}{\mathbb{E} \left ( \left \| SX \right \|^2 \right )} 
\end{equation}
\begin{equation}\label{equation:genprecision}
\hat{P} = \frac{\mathbb{E} \left ( \left \| S\hat{X} - S\hat{X} \right \|^2 \right )}{2 \mathbb{E} \left ( \left \| S\hat{X} \right \|^2 \right )} = \frac{ \mathbb{E}\left [ \left\| S\hat{X} \right\|^2 \right ] - \left\| \mathbb{E} \left [ S\hat{X}\right ] \right\|^2}{\mathbb{E} \left ( \left \| S\hat{X} \right \|^2 \right )}
\end{equation}

where $SX = \log(S[p]X)$, and $S[p]X$ are the combined first and second order coefficients of the MST taken on the signal $X$.

We can also define a separation metric $S$, to measure the distance between the ground truth and generated samples.

\begin{equation}
\label{equation:separation} 
S = \frac{\mathbb{E}  \left ( \left \| SX - S\hat{X} \right \|^2 \right )}{\frac{1}{2}  \left (\mathbb{E}  \left ( \left \| SX \right \|^2  \right ) + \mathbb{E}  \left ( \| S\hat{X} \|^2 \right )  \right )}
\\ = \frac{ \mathbb{E}  \left [  \| SX \|^2 + \| S\hat{X} \|^2\right] - 2 \mathbb{E} [SX]\cdot \mathbb{E} [S \hat{X}]}{ \frac{1}{2}  \left (\mathbb{E}  \left ( \left \| SX \right \|^2  \right ) + \mathbb{E}  \left ( \| S\hat{X} \|^2 \right )  \right )} 
\end{equation}

\begin{table}[htbp!]
\caption{The reconstruction precision and separation for the surface roughness and porosity distributions.}

\begin{tabular}{@{}lllll@{}}
\toprule
Metric     & Porosity (x-axis) & Porosity (y-axis) & Porosity (z-axis) & Surface Roughness \\ \midrule
$P$      & 2.018 $\times 10 ^{-3}$                         & 1.370 $\times 10 ^{-3}$                         & 1.106 $\times 10 ^{-3}$                          &   1.438 $\times 10 ^{-2}$             \\
$\hat{P}$    & 1.422$\times 10 ^{-3}$                      & 1.677 $\times 10 ^{-3}$                     & 9.656$\times 10 ^{-4}$                        &     1.928 $\times 10 ^{-2}$              \\
$S$  & 1.116                         & 1.5794                         & 1.1873                          &           1.905       \\ \bottomrule
\end{tabular}

Upon examination of the values reported (Table \ref{table:mstprecision}), the separation and precision for the porosity distribution are similar to the metrics observed for the surface roughness realizations. This indicates that the newly generated samples are statistically reasonable, as the surface roughness MST metrics are preserved by construction of the microcanonical model, and the metrics for the porosity realizations produced without the MST are on the same order as the surface roughness metrics produced with the MST. 

\label{table:mstprecision}
\end{table}

\FloatBarrier
\section{Conclusion and Next Steps}
\label{sec:conc}
In this work, a method for generating novel stochastic realizations of porosity in AM produced parts is demonstrated. This is carried out by deconstructing the initial tomograph of the part dataset into the composite surface roughness profiles, and the pores comprising the porosity distribution. This enables the construction of new examples in cases with limited amounts of data, as it is able to create new realizations from a single example of a printed part. Following deconstruction, a new surface roughness profile is generated using a Mallat Scattering Transform (MST) based microcanonical model, based on the scattering coefficient properties.  Concurrently, a 3D-DCGAN augments the existing distribution of pores by learning to generate plausible realizations that statistically match the ground truth distribution. Once both components have been created, the conditional spatial distribution of key pore metrics is used to sample from the augmented set of pore examples, creating a new realization of the pore distribution. Finally, the new surface roughness is added to form the boundary of the part. To verify model performance, the generated realizations are compared to the distribution observed in the original data, and match both the univariate distributions of the individual metrics, as well as the bivariate distributions that describe the correlations between the individual metrics. 

This method avoids the computational expense associated with fine-grained probability-based reconstruction methods. Specifically, the efficiency is increased by coarse-graining the resolution of the spatial conditional probability and using a DCGAN for the small-scale recreation of the porosity distribution. Sampling directly using pore metrics also offers a fail-safe to avoid placing outlier pore examples that are not fully resolved. This is necessary, as pore examples that are not fully resolved may be generated when using GAN architectures to create a very large number of pore realizations, as they are complex models with potential failure modes \cite{goodfellow2014generative, arjovsky2017wasserstein}.  The porosity distributions and the surface roughness reconstructions are also quantified by examining the MST coefficients of the generated and ground truth samples, as the MST is a complete metric of multiple point statistics. While the surface roughness exhibits very similar multiple point statistics between the generated and ground truth cases by construction, the method for constructing novel porosity distributions using sampling and GANs is also able to preserve the statistics of the original samples. While the MST works well in the 2D case for generating surface roughness profiles, it is not used to generate the porosity distribution in this work due to the computational requirements of scaling the transform to include a third dimension.

Using a DCGAN to construct individual pores, as opposed to entire part realizations, also avoids constraints in memory that would otherwise scale unfavorably with resolution and part size. Therefore, the proposed model offers increased control over the generation process, when compared to a purely "black-box" GAN based model. In this work, the generation process was applied to create new examples from cylindrical tensile coupons. Therefore, in future studies, this work can be extended by examining performance on larger and more complex part geometries. Additionally, while the resolution of the conditional probability matrices was kept constant in this work, an adaptive resolution that decreases bin size near important features can also minimize the computational expense for generating large parts with small amounts of location-dependent porosity. This work can also act as a more accurate porosity generation method for finite element analysis of simulated additively manufactured samples. The code for this work will be made available upon publication, at the GitHub repository https://github.com/BaratiLab/Porosity-Generator.

\section*{Acknowledgements}
This work was funded by Sandia National Laboratories.
\appendix
\section{Resolution Ablation Study}
\label{sec:sample:appendix}
            
\begin{figure*}[!htbp]
\begin{center}
   \includegraphics[width=1\linewidth]{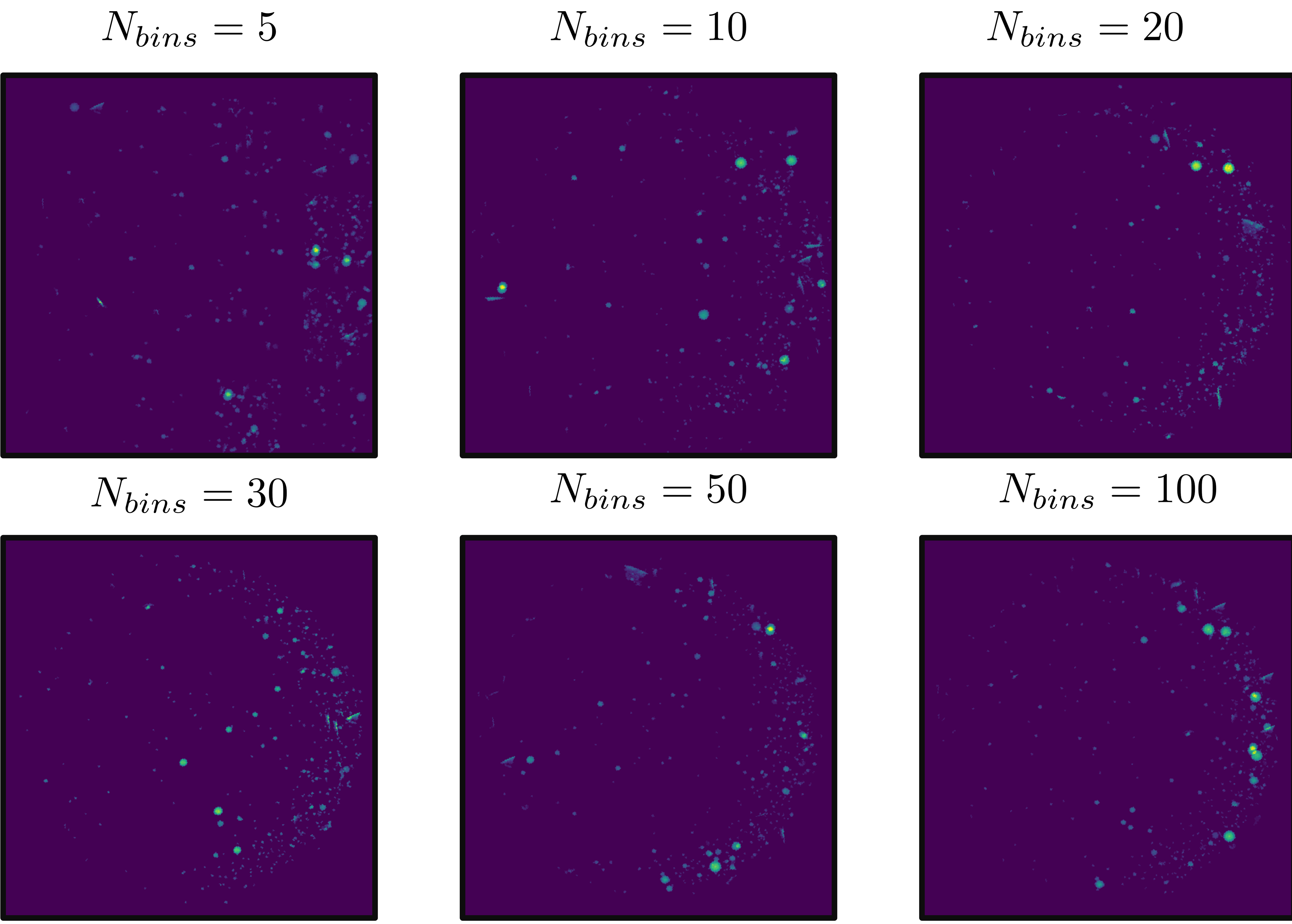}
\end{center}
   \caption{An ablation study performed on the data reveals that increasing resolution results in increasing adherence to the ground truth pore distribution, with a trade off between bin size and computational expense. At $N_{bins} = 20$, the circular shape of the part is resolved. At resolutions higher than 30 bins, the visual difference between the generated samples are marginal.}

\label{ablation}
\end{figure*}

During the reconstruction process, coarse-grained conditional probability matrices of size $\frac{K}{N_{b}} \times \frac{M}{N_{b}} $ are used to sample the probability distributions, to construct an $K\times M \times L $ part realization. A uniform distribution is used to sample at resolutions smaller than the coarse-grained mesh elements. Therefore, $N_{b}$, the number of bins used to traverse each dimension of the x-y cross-section of the part, acts as a hyper-parameter that can be chosen to satisfy the trade-off between computational complexity, and the similarity between the generated and original samples. Figure \ref{ablation} describes the distribution of pores seen as a function of the $N_{b}$ parameter. For very coarse meshes, the reconstruction does not recreate the local structure of the pore distribution, such as the circular boundary that should enclose the 2D distribution of pore examples. However, as $N_{b}$ is increased, the structure of the part becomes increasingly refined. The structure first becomes clearly visible at $N_{b} = 20$, and gradually increases in resolution until $N_{b} = 100$. However, the magnitude of the increased refinement observed decreases at large values of $N_{b}$ as the performance plateaus, indicating that there is an optimum value for $N_{b}$ for reconstruction of the part sample without unnecessary runtime. Here, $N_{b} = 30$ is used for the reconstruction.  

\begin{figure*}[!htbp]
\begin{center}
  \includegraphics[width=1\linewidth]{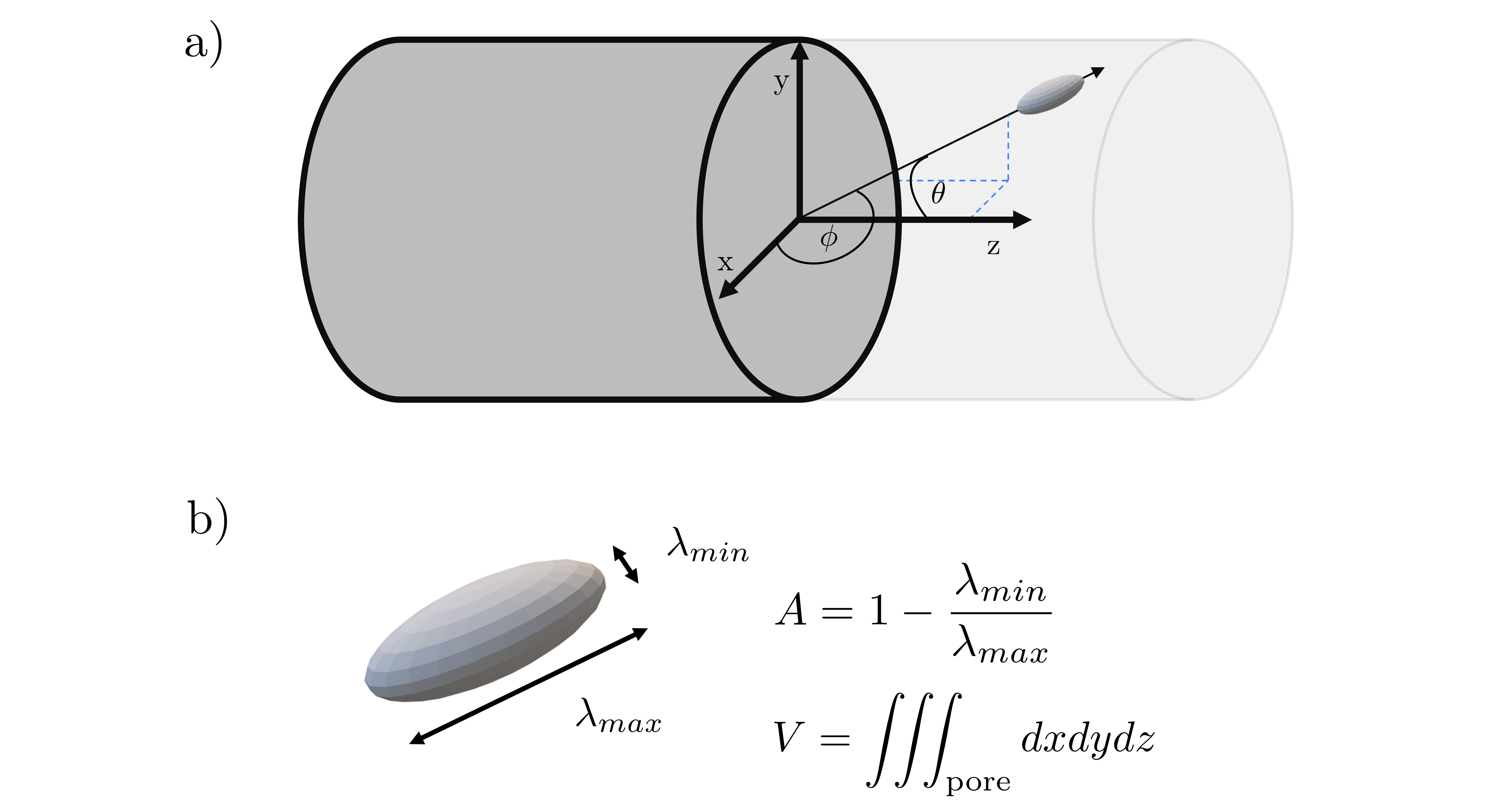}
\end{center}
  \caption{The volume, anisotropy and orientation are extracted from the pore sample for analysis. a) The coordinate system used to define the position of the pore in space, relative to the overall part sample. b) The metrics used to define the shape and size of an individual pore. }

\label{metrics_definition}
\end{figure*}

\section{Hyperparameter and Architecture details}
\label{sec:hyperparams}
\centering
\begin{table}[htbp!]
\caption{Hyperparameters for Generative Adversarial Network training.}
\begin{tabular}{@{}ll@{}}
\toprule
Hyperparameter & Value \\ \midrule
Batch Size & 32 \\
Number of Epochs & 40 \\
Learning Rate & 2E-5 \\
Size of Latent Vector & 100 \\ \bottomrule
\end{tabular}

\end{table}
\begin{table}[htbp!]
\caption{Generator Architecture}
\begin{tabular}{@{}llll@{}}
\toprule
Layer & Type & Kernel Size & Output Channels \\
\midrule
1 & 3D Convolution & 4 x 4 & 512 \\
2 & 3D Convolution & 4 x 4 & 256 \\
3 & 3D Convolution & 4 x 4 & 128 \\
4 & 3D Convolution & 4 x 4 & 64 \\
5 & 3D Convolution & 4 x 4 & 2\\ \bottomrule
\end{tabular}

\end{table}

\begin{table}[htbp!]
\caption{Discriminator Architecture}
\begin{tabular}{@{}llll@{}}
\toprule
Layer & Type & Kernel Size & Output Channels \\
 \midrule
1 & 3D Convolution & 4 x 4 & 16 \\
2 & 3D Convolution & 4 x 4 & 32 \\
3 & 3D Convolution & 4 x 4 & 64 \\
4 & 3D Convolution & 4 x 4 & 128 \\ 
5 & 3D Convolution & 4 x 4 & 1 \\ \bottomrule
\end{tabular}

\end{table}

 \bibliographystyle{elsarticle-num} 
 \bibliography{pore_gan_references}





\end{document}